\documentclass[lettersize,journal]{IEEEtran}
\usepackage{amsmath,amsfonts}
\def\BibTeX{{\rm B\kern-.05em{\sc i\kern-.025em b}\kern-.08em
    T\kern-.1667em\lower.7ex\hbox{E}\kern-.125emX}}
\usepackage{balance}

\usepackage[numbers]{natbib}
\usepackage{multicol}
\usepackage[bookmarks=true]{hyperref}
\usepackage{graphics} 
\usepackage{epsfig} 
\usepackage{times} 
\usepackage{amssymb}  
\usepackage{hyperref} 
\usepackage{xcolor} 
\usepackage{arydshln} 
\usepackage{color}
\usepackage{float}
\usepackage{multirow}
\usepackage{booktabs}
\usepackage{tikz}
\usepackage{wrapfig}
\usepackage{units}
\usepackage{bm}
\usepackage{tabularx}
\usetikzlibrary{calc ,math, positioning, intersections, decorations, external, plotmarks, shapes}
\usepackage{pgfplots}
\usepackage{pgfplotstable}
\usepgfplotslibrary{statistics}
\pgfplotsset{compat=1.16}
\usepackage[skip=0pt,font=footnotesize]{caption}
\usepackage{subcaption}

\newcommand{\comment}[1]{{\color{black} #1}}

\newcolumntype{C}{>{\centering\arraybackslash}X}
\newcolumntype{P}[1]{>{\centering\arraybackslash}p{#1}}
\definecolor{textblue}{HTML}{0072B2}
\definecolor{textorange}{HTML}{D55E00}
\definecolor{textgreen}{HTML}{008060}

\def\orb{\mbox{ORB-SLAM3}}
\def\orbf{\mbox{ORB-SLAM3-OF}}
\def\orbfu{\mbox{ORB-SLAM3-OF-U}}
\def\orbfTable{ORB-SLAM3-OF} 
\def\orbfuTable{ORB-SLAM3-OF-U} 
\def\droidvo{DROID-VO}
\def\droidslam{DROID-SLAM}
\def\dpvo{DPVO}
\def\dpvslam{DPV-SLAM}
\def\vslam{V-SLAM}
\def\tartanair{TartanAir}
\def\uzhfpv{UZH-FPV}
\def\atet{$\text{ATE}_\text{T}$}
\def\ater{$\text{ATE}_\text{R}$}
\def\ate{$\text{ATE}$}
\def\qf{\textit{Is learned 2D data association a key enabler of deep learning-based systems' superior performance?}}
\def\qs{\textit{Does the key lie in combining learned 2D data association with learned uncertainty?}}
\def\qt{\textit{Or is the recurrent architecture the crucial mechanism that brings these components together to outperform classical methods?}}

\def\question{\textit{Is the superior performance of deep learning-based systems driven primarily by learned 2D data association, the combination of learned 2D data association and uncertainty, or the recurrent architecture itself?}}

\title{Why does Deep Learning Improve Visual SLAM?}

\begin{document}

\author{Giovanni Cioffi and Davide Scaramuzza
\thanks{The authors are with the Robotics and Perception Group, Department of Informatics, University of Zurich, Switzerland, \protect\url{https://rpg.ifi.uzh.ch}.\newline
This work was supported by the European Union’s Horizon Europe Research and Innovation Programme under grant agreement No. 101120732
(AUTOASSESS) and the European Research Council (ERC) under grant agreement No. 864042 (AGILEFLIGHT).}}

\maketitle

\begin{abstract}
Visual SLAM is a well-established technology utilized in a wide range of real-world applications.
However, its performance still degrades under challenging visual conditions, such as low texture, severe motion blur, and poor illumination. 
Systems based on deep learning outperform classical geometry-based ones and achieve state-of-the-art results by combining learned 2D data association and uncertainty with differentiable geometric optimization \comment{in recurrent architectures}.
Still, it remains unclear exactly which components are fundamentally responsible for this success. 
In this paper, we ask: \comment{\question}\
We investigate this question empirically by conducting a controlled study.
Our findings reveal that the success of DL-based V-SLAM systems hinges on learned 2D data association and uncertainty rather than their recurrent architecture, underscoring the necessity of learning-based paradigms for the design of these components.
%
\end{abstract}


\section*{Supplementary Material}\label{sec:SupplementaryMaterial}

\textbf{Video}: \url{https://youtu.be/EiuZ7MT0iVc}

\textbf{Code}:\url{https://github.com/uzh-rpg/why_does_deep_learning_improve_visual_slam}
\section{Introduction}\label{sec:Introduction}
\IEEEPARstart{V}{isual} Simultaneous Localization and Mapping (\vslam) estimates the motion of a camera while reconstructing a map of the surrounding environment.
It is a key enabling technology for robot navigation~\cite{ebadi2024present}, autonomous driving~\cite{janai2020computer}, augmented and virtual reality (AR/VR)~\cite{engel2023project}, assistive systems for visually impaired individuals~\cite{pfreundschuh2025sight}, and medical procedures~\cite{most2025acquiring}.

For nearly three decades, \vslam\ systems have followed a frontend--backend architecture~\cite{slam-handbook} rooted in geometry-based computer vision.
In feature-based methods, the frontend detects and tracks visual measurements through handcrafted keypoints and descriptors, while the backend estimates camera poses and 3D structure by minimizing reprojection errors through bundle adjustment (BA).
Alternatively, direct methods bypass explicit feature extraction and operate directly on pixel intensities, and the backend minimizes photometric errors.
While these systems achieve high accuracy in many situations~\cite{slam-handbook}, they exhibit inherent fragilities in challenging real-world conditions. 
Specifically, the reliance of feature-based methods on repeatable features and descriptors, and the assumption of photometric consistency in direct methods, often lead to failure in very low-texture scenes~\cite{wang2020tartanair}, severe motion blur~\cite{delmerico2019we}, high dynamic range scenarios~\cite{gehrig2021dsec}, and environments with poor illumination~\cite{helmberger2022hilti} or dynamic agents~\cite{palazzolo2019iros}.

\begin{figure}[t!]
    \centering
    \includegraphics[width=1.0\linewidth]{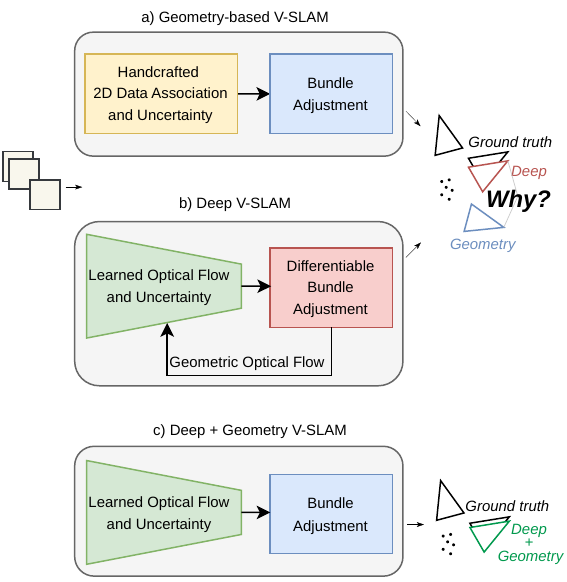}
    \vspace{-1mm}
    \caption{Uncovering why deep learning improves visual SLAM. (a) Classical geometry-based Visual SLAM relies on handcrafted feature matching and geometric optimization. (b) Modern deep SLAM systems achieve superior robustness, but it is unclear which learned components are responsible.
    \comment{(c) By integrating learned 2D data association and uncertainty into a classical geometry-based pipeline, we isolate their impact and demonstrate that these two components enable a classical system to achieve state-of-the-art results.}}
    \label{fig:figure1}
\end{figure}

Deep learning (DL) has changed the performance landscape of \vslam.
Early end-to-end methods directly regressed camera motion from image sequences~\cite{wang2017deepvo,wang2021tartanvo}, but often struggled to generalize beyond the visual conditions and motion distributions seen during training.
In contrast, the most successful systems adopt a hybrid paradigm: they combine learned visual frontends with geometric optimization backends in \comment{recurrent architectures}~\cite{teed2021droid,teed2022deep,lipson2024deep}.
A representative example is~\droidslam~\cite{teed2021droid}, which couples a RAFT-based optical-flow network~\cite{teed2020raft} with geometric bundle adjustment.
The optical flow network predicts dense correspondences and confidence estimates that define reprojection constraints for bundle adjustment. 
Bundle adjustment then estimates camera poses and depths of the 3D map points, which are used to compute an updated geometric optical flow that is fed back into the network to guide the next iteration.
This recurrent architecture is a key innovation contributing to the strong performance of \droidslam.
\droidslam\ is a dense method that tracks all pixels, albeit at significantly reduced image resolution.
Building on this paradigm, subsequent work such as \dpvo\ and \dpvslam, demonstrates that comparable accuracy can be achieved with a sparse representation consisting of randomly selected image patches.
Despite their structural differences, these systems share a common reliance on learned optical flow, uncertainty estimation, and a recurrent architecture.
\comment{\droidslam, \dpvo, and \dpvslam\ define the state-of-the-art in trajectory estimation, and they are our chosen systems for this study.
While more recent systems leveraging novel scene representations~\cite{zhu2022nice, zhu2024nicer, matsuki2024gaussian, yugay2023gaussian} and 3D foundation models~\cite{murai2025mast3r, maggio2025vggt-slam, maggio2025vggt-slam2} achieve remarkable accuracy in the dense reconstruction of indoor environments, they lack robustness in unstructured outdoor environments and under high-speed camera motions.}

\comment{The success of DL-based \vslam\ raises a critical question for the field: \question}\
Answering this question is essential: it determines whether future \vslam\ systems should abandon traditional handcrafted visual frontends entirely, applying geometric principles solely in the backend optimization, and retain a feed-forward architecture.

In this paper, we provide a controlled empirical answer to this question.
We take \orb, a mature and highly optimized geometry-based \vslam\ system, and replace its descriptor-based data association module with the optical-flow network used in \droidslam.
This yields \orb\ with learned optical flow (\orbf), a system that preserves the classical SLAM backbone of \orb\ while using learned correspondences in place of handcrafted descriptor matches.
Additionally, in \orb\ with learned optical flow and uncertainty (\orbfu), we utilize the uncertainty estimates predicted by the optical-flow network to weight the residuals in the bundle adjustment.
This allows us to isolate and quantify the effect of learned uncertainty, which is used by \droidslam, \dpvo, and \dpvslam\ to weight geometric residuals, \comment{and has recently been applied to improve stereo matching in visual odometry~\cite{qiu2025mac} and multi-sensor fusion~\cite{zhao2025resilient} in challenging conditions}.
Our approach changes only the data association and uncertainty mechanisms within the same SLAM pipeline. 
Consequently, we can directly measure how much of the performance gap between geometry- and learning-based \vslam\ is explained by these two components.

We evaluate the standard \orb\ pipeline, \orbf, and \orbfu\ against the leading learning-based systems, including \droidslam, \dpvo, and \dpvslam.
Experiments are conducted on \tartanair~\cite{wang2020tartanair} and \uzhfpv~\cite{delmerico2019we}, the two most challenging benchmarks that expose \vslam\ systems to diverse motion patterns and adverse perceptual conditions, including low texture, poor illumination, and severe motion blur.
Our results show that replacing handcrafted descriptor matching with learned optical flow substantially improves \orb.
Moreover, adding learned uncertainty further improves robustness in visually degraded scenarios, where correspondence estimation is difficult and uniform residual weighting becomes unreliable.
\comment{The resulting \orbfu\ achieves state-of-the-art performance, rivaling modern deep \vslam\ systems on in-distribution data (TartanAir), and remarkably outperforming them on out-of-distribution sequences (UZH-FPV).}


\comment{Our findings demonstrate that a classical feed-forward SLAM system equipped with learned 2D data association and uncertainty yields optimal performance. 
This emphasizes a key insight: the success of DL-based \vslam\ systems hinges on learned 2D data association and uncertainty rather than their recurrent architecture, underscoring the necessity of learning-based paradigms for the design of these components.}
We believe that the findings of this work will guide the design of the next generation of \vslam\ systems.

To summarize, as illustrated in Fig.~\ref{fig:figure1}, our main contributions are:
\begin{itemize}
    \item A controlled framework for studying learned data association in \vslam. 
    By replacing the handcrafted descriptor-based matching module of \orb\ with a learned optical-flow correspondence estimator derived from \droidslam, we isolate the effect of learned data association while keeping the classical \orb\ pipeline fixed.
    \item A systematic evaluation of learned uncertainty for weighting residuals in the bundle adjustment.
    We incorporate uncertainty estimates predicted by the optical-flow network into the bundle adjustment of \orb\ and quantify their impact on pose estimation.
    \item An extensive empirical study on the most challenging \vslam\ benchmarks.
    Through experiments on \tartanair\ and \uzhfpv, we show that learned correspondences and learned uncertainty substantially improve a classical \vslam\ system, enabling \orbfu\ to achieve state-of-the-art performance.
\end{itemize}
\section{Related Work}\label{sec:Related_Work}
The field of \vslam\ has undergone a profound evolution, transitioning from early probabilistic formulations to modern spatial intelligence systems. 
A comprehensive overview of this progression is provided in the recently published \textit{SLAM Handbook}~\cite{slam-handbook}. 
The volume characterizes SLAM not merely as a state estimation problem but as a multidisciplinary challenge involving factor graphs~\cite{dellaert2012factor}, Lie groups~\cite{sola2018micro}, and robust optimization~\cite{grisetti2011tutorial}. 
The current research frontier in \vslam\ is defined by the synergy between classical geometric and deep learning methods.

\subsection{Classical Geometry-Based SLAM Systems}
Classical geometry-based \vslam\ systems are generally categorized by their frontend methodology into \textit{feature-based} and \textit{direct} methods. 
\textit{Feature-based} approaches, such as the ORB-SLAM series~\cite{mur2015orb, mur2017orb, campos2021orb}, rely on the extraction and matching of handcrafted keypoints~\cite{rosten2006machine} and descriptors~\cite{rublee2011orb} to estimate camera motion and scene structure by minimizing reprojection errors.
In contrast, direct methods like DSO~\cite{engel2017direct} operate directly on pixel intensities, minimizing photometric errors. 
While direct methods are capable of generating denser maps and performing better in low-texture environments, they are significantly more sensitive to illumination changes, rolling shutter artifacts, and non-Lambertian surfaces.
Hybrid methods exist, such as SVO~\cite{forster2016svo}, which extract handcrafted keypoints and minimize photometric errors.

The backend architectures of classical systems are broadly categorized into \textit{filter-based} and \textit{optimization-based} frameworks. 
Early systems like MonoSLAM~\cite{davison2007monoslam} and the computationally efficient MSCKF~\cite{mourikis2007multi} utilized Extended Kalman Filters (EKFs) to maintain a state vector. 
However, as noted in~\cite{slam-handbook}, the field has largely shifted toward non-linear least squares optimization over factor graphs. 
Optimization-based systems, such as OKVIS~\cite{leutenegger2015keyframe} and VINS-Mono~\cite{qin2018vins}, allow for the re-processing of past states through bundle adjustment, offering superior consistency and accuracy~\cite{strasdat2012visual}.

The primary advantage of classical geometry-based SLAM lies in its mathematical explainability and efficiency on resource-constrained hardware. 
However, these systems often struggle in scenarios involving aggressive motion, severe motion blur, or extreme lighting conditions.
For a comprehensive survey of these classical failure modes, we refer the reader to~\cite{sh-ch7-vision}.

\subsection{Deep Learning-Based SLAM Systems}
The advent of Deep Learning has introduced new paradigms to address the limitations of classical geometry. 
Early \textit{end-to-end} systems, such as DeepVO~\cite{wang2017deepvo} and TartanVO~\cite{wang2021tartanvo}, attempted to regress camera poses directly from image sequences. 
While these methods exhibit robustness to visual artifacts, they struggle with generalization to unseen environments.

A successful paradigm involves \textit{hybrid} architectures that integrate learned frontends with differentiable geometric backends~\cite{teed2021droid, teed2022deep, lipson2024deep}. 
D3VO~\cite{yang2020d3vo} was one of the first systems to leverage deep learning for robust data association while maintaining a geometric optimization, although it did not utilize a differentiable backend.
\droidslam~\cite{teed2021droid} pioneered the hybrid approach by combining a recurrent optical flow network with a differentiable BA layer. 
By utilizing a bundle adjustment module, the system enforces geometric consistency while benefiting from the robust data association of the flow network. 
Subsequent works like \dpvo~\cite{teed2022deep} and \dpvslam~\cite{lipson2024deep} have further refined this by employing sparse patch-based representations, significantly reducing computational overhead while maintaining state-of-the-art accuracy.

The most recent evolution in the field is the emergence of \textit{3D foundation models} and \textit{neural representations}. 
Systems like iMAP\cite{sucar2021imap} and NICE-SLAM~\cite{zhu2022nice, zhu2024nicer} utilize Neural Radiance Fields (NeRF)~\cite{mildenhall2021nerf} for joint tracking and dense mapping.
More recently, Gaussian Splatting SLAM (GS-SLAM)~\cite{matsuki2024gaussian, yugay2023gaussian} has gained traction for its ability to provide high-fidelity, real-time rendering. 
Furthermore, systems such as MASt3r-SLAM~\cite{murai2025mast3r} and VGGT-SLAM~\cite{maggio2025vggt-slam, maggio2025vggt-slam2} build upon feed-forward 3D geometric foundation models such as MASt3r~\cite{leroy2024grounding} and VGGT~\cite{wang2025vggt}, which have enabled dense 3D reconstruction from uncalibrated image pairs in a single pass. 
These models leverage large-scale transformer pre-training to resolve projective ambiguity and establish dense correspondences even under extremely low visual overlap.

The advantage of DL-based systems is their robustness to perceptual challenges. 
Our work specifically bridges these two worlds by investigating how integrating the robust data association and uncertainty estimation of DL-based systems into the reliable framework of classical geometry-based \vslam\ impacts the system's performance.
\section{Methodology}\label{sec:Methodology}
This section presents the framework designed to evaluate the impact of learned 2D data association and uncertainty within a classical geometry-based \vslam\ architecture. 
To address our research question, we systematically replace the data association module of a state-of-the-art classical baseline with a DL-based optical-flow estimator. 
This modular approach allows for a direct comparison between the two paradigms while maintaining a geometric bundle adjustment backend for pose optimization.
The remainder of this section is organized as follows: first, we provide the rationale for the selection of the \vslam\ systems used in our study. 
Second, we describe the core architectural components of these systems. 
Finally, we provide a detailed description of the methodology used to integrate the learning-based 2D data association and uncertainty module into the classical \vslam\ pipeline.

\begin{figure}[t!]
    \centering
    \includegraphics[width=1.0\linewidth]{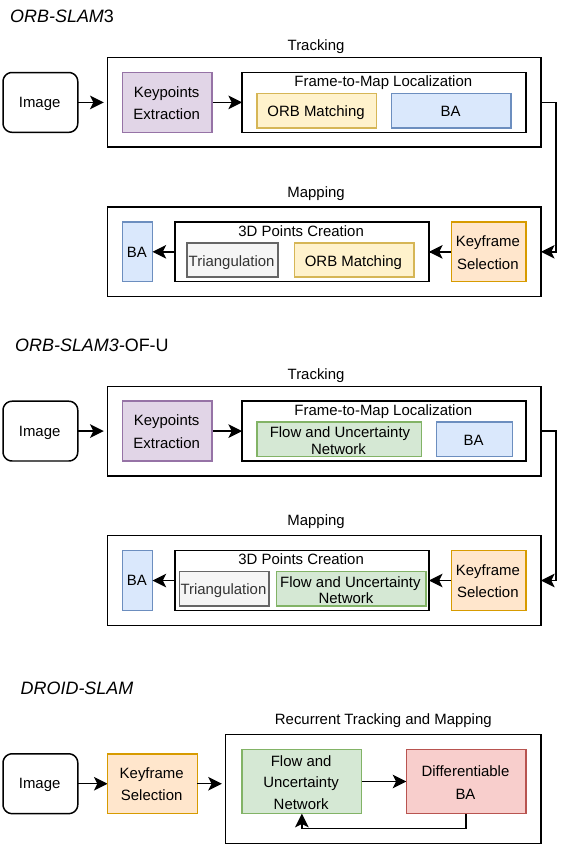}
    \vspace{-1mm}
    \caption{Diagram of the classical \orb\ architecture (top), \orb\ with integrated learned optical flow and uncertainty (middle), and the architecture common to \droidslam, \dpvo, and \dpvslam\ (bottom).}
    \label{fig:flow_charts}
\end{figure}

\subsection{Choice of \vslam\ Systems}
The selection of representative systems is critical for the validity of this empirical study. 
While numerous geometry- and DL-based \vslam\ systems demonstrate high performance under specific environmental or sensor configurations, we prioritize systems that have shown consistent state-of-the-art performance across diverse scenarios. 
The frameworks selected for this study are characterized by their mature, consolidated codebases and their status as established benchmarks within the robotics community.

For the geometry-based system, we utilize \orb\ (specifically the ORB-SLAM3 implementation). 
\orb\ is widely regarded as the most robust geometry-based \vslam\ system, employing a comprehensive pipeline that includes tracking, local mapping, and global loop closure. 
Its 2D data association relies on descriptor matching, a methodology representative of the traditional SLAM paradigm. 
Replacing descriptor matching with a learned alternative allows us to quantify the benefits of DL-based data association against a highly optimized classical method.

For the DL-based systems, we evaluate \droidslam, \dpvo, and \dpvslam. 
\droidslam\ introduced a recurrent architecture combining a learned frontend with a differentiable bundle adjustment layer, demonstrating superior accuracy and generalization over end-to-end approaches. 
Its follow-up works, \dpvo\ and \dpvslam, extend this concept by employing a sparse patch-based representation to improve computational efficiency. 
These systems currently define the state-of-the-art in trajectory estimation accuracy.
While emerging methods leveraging 3D foundation models and neural representations have demonstrated remarkable accuracy and dense reconstruction capabilities in indoor settings, they often lack the robustness required for high-velocity maneuvers or unstructured outdoor environments.

\subsection{Learning-Based \vslam\ Systems}
\droidslam\ operates on a frame graph where edges represent co-visibility between images. 
For each edge, the architecture extracts feature maps and constructs correlation volumes to evaluate visual similarity.
The optical flow network of \droidslam\ is a convolutional GRU-based update operator that iteratively predicts flow revisions and associated confidence weights, which are then passed to a differentiable dense BA layer that refines camera poses and pixel-wise depth.
The refined camera poses and depths are used to compute the new flow prediction that is fed as one of the inputs to the flow network.
The recurrent structure of \droidslam\ is one of the reasons for its success.
\droidslam\ concludes with a bundle adjustment optimizing the entire trajectory of keyframes and 3D map points.
In our experiments, we also evaluate a version, \droidvo, where this bundle adjustment is disabled.

The reliance of \droidslam\ on dense flow maps results in high computational and memory costs.
To address the efficiency bottleneck, \dpvo\ was proposed as a sparse analog to \droidslam.
Instead of predicting dense flow for every pixel, \dpvo\ tracks a sparse collection of image patches across time. 
These patches are sampled randomly, a strategy found to be surprisingly effective compared to traditional keypoint detectors.
\dpvo\ utilizes a bipartite patch graph to represent the relationship between patches and frames. 
Its recurrent update operator includes temporal convolutions and global message-passing layers to propagate information between edges in the patch graph. 
\dpvo\ runs faster and requires lower memory than \droidslam\ while maintaining competitive accuracy.

\dpvslam\ extends \dpvo\ into a complete SLAM framework by introducing efficient loop closure and global optimization.
A major challenge in deep SLAM is the sequential nature of CUDA operations, which often causes frame rate drops during backend optimization. 
\dpvslam\ addresses this by maintaining a single, shared patch graph that incorporates both local odometry factors and long-range loop closure factors.
The system employs two loop closure mechanisms: proximity loop closure and classical loop closure.
Proximity loop closure uses the predicted flow and camera poses to detect loops and inserts long-range proximity factors into the patch graph for global bundle adjustment. 
Classical loop closure utilizes image retrieval (e.g., BoW2~\cite{galvez2012bags}) and pose graph optimization on the CPU to correct for significant scale drift.
In this work, we enable the proximity loop closure but disable the classical loop closure module.
\dpvo\ and \dpvslam\ use the same recurrent flow network architecture as \droidslam.
The shared architecture common to \droidslam, \dpvo, and \dpvslam\ is illustrated in the bottom diagram of Fig.~\ref{fig:flow_charts}.

\subsection{Integrating Learned Data Association in \orb}
\orb\ is a feature-based \vslam\ system designed for real-time operation in diverse indoor and outdoor environments. 
It uses FAST corners as keypoints and ORB as descriptors. 
Keypoints and corresponding descriptors are extracted on every incoming frame.
The system's architecture is built around three main parallel threads: Tracking, Local Mapping, and Loop Closing. 
ORB features are used for all tasks (tracking, mapping, relocalization, and loop detection), which ensures efficiency.
The tracking thread localizes the camera for every frame by matching ORB descriptors to a local map and optimizing the camera pose using motion-only BA.
The local mapping thread processes new keyframes and optimizes the local reconstruction through local BA. 
Keyframes are spawned generously to maintain tracking robustness during exploration, but are later subject to a restrictive culling policy. 
This removes redundant information, resulting in a compact and trackable map that enables lifelong operation.
The loop closing thread searches for loops with every new keyframe. 
When a loop is detected, the system computes a 7-DoF similarity transformation to correct accumulated scale drift and performs a pose-graph optimization over an essential graph. 
This sparse subgraph maintains global consistency while remaining computationally efficient for large-scale environments.
For monocular cameras, \orb\ utilizes an automatic initialization procedure based on model selection. 
It computes two parallel models, a homography for planar scenes and a fundamental matrix for non-planar scenes, and uses a heuristic to select the most appropriate one, ensuring a reliable initial map is created only when sufficient parallax is present.
While the original version was strictly monocular, subsequent iterations up to ORB-SLAM3 have expanded the system's capabilities to support stereo cameras, RGB-D sensors, and Inertial Measurement Units (IMUs).

To isolate the contribution of learned data association, our integration ensures that the fundamental system design of \orb\ (tracking, mapping, and loop closing threads) remains unchanged. 
We replace only the 2D correspondence engine with the optical flow network from \droidslam.
The system architecture of \orb\ is illustrated in the top diagram of Fig.~\ref{fig:flow_charts}.
In this work, we utilize the monocular mode.

\subsubsection{Flow Network Inference}
The flow network, taken from \droidslam, is pre-trained on the \tartanair\ dataset. 
During inference, input images are converted to grayscale and downsampled to a resolution of $80 \times 60$ (1/8th the original \tartanair\ resolution). 
A feature network computes feature maps, from which a 4D correlation volume is derived. 
Context features are also extracted by the context network.
The core of the inference is a $3 \times 3$ convolutional GRU (ConvGRU) that iteratively refines flow estimates.

Unlike the standard \droidslam\ implementation, we initialize the process with a null-flow prior to remove any reliance on a motion model prior. 
Consequently, the correspondence field used to extract features from the correlation volume is the identity.
The ConvGRU takes as inputs the correlation features, the induced optical flow (the difference between the current correspondence and the initial null-flow prior), and the context features.
It outputs flow revisions, namely a revision to the current correspondence field and a confidence map.
The ConvGRU iterates until the mean flow revision falls below 0.0625 (equivalent to approximately 0.5 pixels at original image resolution) or a maximum of 20 iterations is reached. 
The final flow estimates are upsampled to the original image resolution using bilinear interpolation.

\subsubsection{Feature Matching Workflow}
In the standard \orb\ tracking thread, 3D map points associated with keypoints from the previous frame (or keyframe) are reprojected onto the current frame using a constant-velocity motion model. 
Matches are then sought by comparing ORB descriptors within a local search window around the reprojection.
These matches are the inputs to a motion-only BA that optimizes the pose of the current camera frame.
If frame-to-frame tracking fails, namely, the number of inliers in the BA is below a threshold (set to 7 in our experiments), the system attempts frame-to-keyframe localization. 
If this localization also fails, the system enters a lost state and attempts to localize the current camera on the global map using BoW2.

In our proposed \orbf\ and \orbfu\ variants, we replace the descriptor matching module. 
Given a 3D map point associated with a 2D keypoint on a source frame $(\text{u}_s, \text{v}_s)$, we determine its predicted location on the target frame ($\text{u}_t, \text{v}_t$) by using the network-predicted optical flow $(\text{f}_u, \text{f}_v)$: $(\text{u}_t, \text{v}_t) = (\text{u}_s + \text{f}_u, \text{v}_s + \text{f}_v)$. 
We then establish a correspondence with the closest feature within a 2-pixel radius of $(\text{u}_t, \text{v}_t)$. 
If no feature exists within this radius, the map point is marked as not visible in the current frame. 
This process is applied both for frame-to-frame tracking and frame-to-keyframe localization.

\begin{table*}[t!]
\caption{Absolute Trajectory Error, in translation (\atet) and rotation (\ater), on the \tartanair\ dataset. Lower values indicate better performance, \textbf{bold} entries denote the best results in translation and rotation. The symbol '-' denotes that the SLAM system failed to maintain tracking for the entire sequence. \textcolor{textblue}{The learned \orb\ versions outperform the standard version of \orb.} \textcolor{textgreen}{The performance of the two variants with and without learned uncertainty is generally comparable.} \textcolor{textorange}{The performance of the learned \orb\ versions is on par with that of the DL systems.}}
\vspace{3pt}
\label{tab:tartanair}
\setlength{\tabcolsep}{8pt}
\begin{tabularx}{\textwidth}{C|C|C|C|C|C|C|C}
\toprule 
\multirow[c]{1}{=}[-12pt]{\centering Sequence} & \multicolumn{7}{c}{Evaluation Metric: \atet [\unit{m}] / \ater [\unit{deg}]}  \\[4pt]
& \orb & \orbfTable & \orbfuTable & \droidvo & \droidslam & \dpvo & \dpvslam \\ [4pt]
 \midrule
 ME000 & 0.17 / 0.85 & \textcolor{textgreen}{0.27 / 0.49} & \textcolor{textgreen}{0.20 / 0.62} & 0.14 / 0.47 & \textbf{0.10} / \textbf{0.20} & 0.13 / 0.36 & 4.98 / 20.08\\
 ME001 & 0.07 / 0.24 & \textcolor{textgreen}{0.06 / 0.24} & \textcolor{textgreen}{0.07 / 0.23} & 0.09 / 0.18 & 0.05 / \textbf{0.10} & \textbf{0.04} / 0.14 & 0.06 / 0.12\\
 ME002 & 1.84 / 1.65 & \textcolor{textgreen}{0.15 / 0.35} & \textcolor{textgreen}{0.24 / 0.41} & 0.14 / 1.31 & 0.13 / 0.32 & \textbf{0.11} / \textbf{0.15} & 1.49 / 1.51\\
 ME003 & 2.42 / 4.45 & \textcolor{textgreen}{1.11 / \textbf{0.41}} & \textcolor{textgreen}{0.69 / 0.62} & 1.17 / 1.37 & 1.15 / 1.28 & \textbf{0.41} / 0.45 & 1.31 / 5.44\\
 ME004 & - & \textcolor{textblue}{1.79 / 3.95} & \textcolor{textblue}{1.20 / 3.39} & 0.84 / 0.76 & 0.90 / 0.90 & \textbf{0.42} / \textbf{0.59} & 0.52 / 0.88\\
 ME005 & 0.10 / 0.91 & \textcolor{textgreen}{0.10 / 0.91} & \textcolor{textgreen}{\textbf{0.09} / 1.08} & \textbf{0.09} / 0.66 & 0.10 / 0.59 & 0.13 / \textbf{0.33} & 0.21 / 0.95\\
 ME006 & - & \textcolor{textblue}{0.61 / 1.14} & \textcolor{textblue}{0.57 / 0.64} & 0.57 / 0.92 & \textbf{0.45} / 1.10 & 1.00 / \textbf{0.58} & 9.22 / 2.50\\
 ME007 &\textbf{ 0.05} / 0.28 & \textcolor{textgreen}{0.10 / 0.43} & \textcolor{textgreen}{0.08 / 0.36} & 0.24 / 0.79 & 0.08 / \textbf{0.25} & 0.19 / 0.66 & 1.06 / 5.69\\
 MH000 & 1.17 / 2.82 & \textcolor{textgreen}{0.64 / 0.96} & \textcolor{textgreen}{0.51 / 1.28} & 0.31 / 0.62 & \textbf{0.07} / \textbf{0.26} & 0.52 / 1.41 & 0.53 / 1.62\\
 MH001 & 0.09 / 1.30 & \textcolor{textgreen}{0.06 / 0.69} & \textcolor{textgreen}{\textbf{0.04} / 0.46} & 0.44 / 1.01 & 0.39 / 0.75 & \textbf{0.04} / 0.49 & \textbf{0.04} / \textbf{0.42}\\
 MH002 & - & \textcolor{textblue}{0.09 / 0.40} & \textcolor{textblue}{0.06 / 0.33} & 0.07 / 0.11 & \textbf{0.02} / \textbf{0.06} & 0.03 / 0.11 & 1.34 / 4.66\\
 MH003 & - & \textcolor{textblue}{0.08 / 0.82} & \textcolor{textblue}{0.11 / 0.70} & 0.05 / 0.50 & \textbf{0.01} / \textbf{0.20} & 0.08 / 0.25 & 0.04 / 0.28\\
 MH005 & - & \textcolor{textblue}{0.98 / 1.39} & \textcolor{textblue}{0.45 / 0.89} & 0.53 / 0.79 & 0.52 / 0.77 & \textbf{0.16} / \textbf{0.26} & 0.65 / 1.96\\
 MH006 & - & \textcolor{textblue}{0.28 / 1.35} & \textcolor{textblue}{0.24 / 1.08} & 0.16 / 0.40 & 0.17 / 0.51 & \textbf{0.08} / \textbf{0.29} & 0.48 / 3.82\\
 MH007 & - & \textcolor{textblue}{0.86 / 1.64} & \textcolor{textblue}{0.61 / 0.96} & 0.83 / 0.33 & \textbf{0.09} / \textbf{0.16} & 0.25 / 0.28 & 4.51 / 5.79\\
 \midrule
 Average & - & \textcolor{textorange}{0.48 / 1.01} & \textcolor{textorange}{0.34 / 0.87} & \textcolor{textorange}{0.38 / 0.61} & \textcolor{textorange}{0.28 / 0.50} & \textcolor{textorange}{\textbf{0.24} / \textbf{0.42}} & 1.76 / 3.72\\
\bottomrule
\end{tabularx}
\end{table*}
\begin{figure*}[t!]
    \centering
    \includegraphics[width=1.0\linewidth]{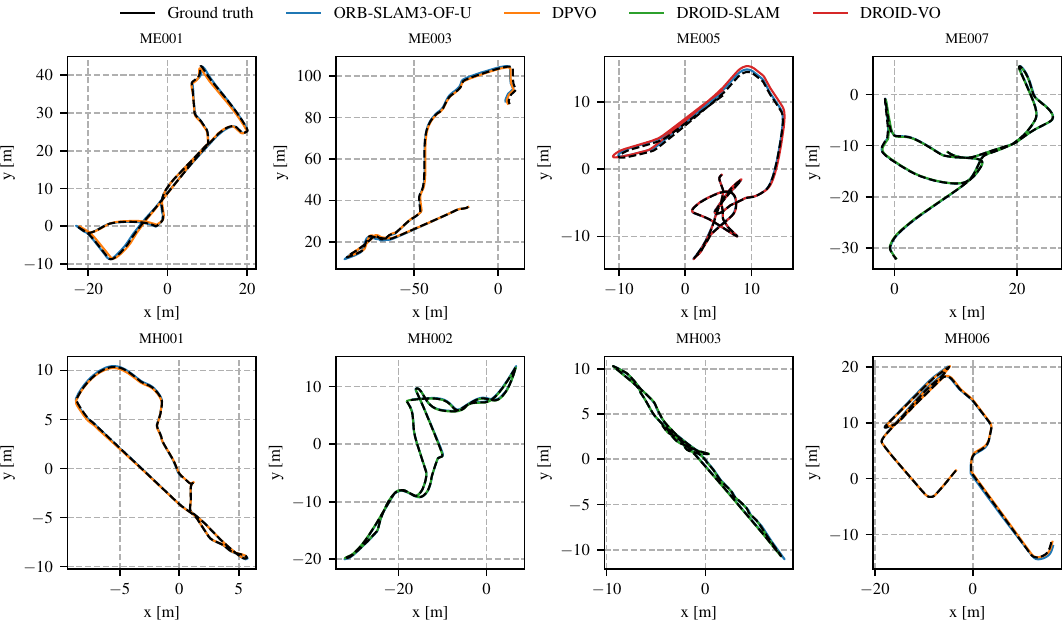}
    \vspace{-1mm}
    \caption{Top-down trajectories for eight sequences from the \tartanair\ dataset. Ground truth is compared against trajectories from \orbfu\ and the best-performing DL-based SLAM method according to \atet. All predicted trajectories are aligned to the ground truth using the method~\cite{umeyama2002least} based on the first 100 frames of each sequence.}
    \label{fig:traj_topview_viz_tartanair}
\end{figure*}

Additionally, we utilize the learned optical flow network to establish feature correspondences between keyframes within the local mapping thread, provided the keyframe pairs reside within a 5-frame temporal window. 
For temporal gaps exceeding this threshold, the system reverts to the standard BoW-based matching approach. 
The underlying matching mechanism follows the same process as the one used for frame-to-frame and frame-to-keyframe tracking.
In the \orbf\ variant, a match is accepted if it satisfies the geometric and distance criteria previously described for the tracking modules. 
In the \orbfu\ variant, we impose an additional validity requirement, where a candidate match is only considered valid if its predicted flow confidence exceeds a threshold of 0.3. 
These established feature matches are subsequently employed for the triangulation of new 3D map points.
The system architecture of \orbf\ and \orbfu\ is illustrated in the middle diagram of Fig.~\ref{fig:flow_charts}.

\subsubsection{Learned Uncertainty in Bundle Adjustment}
In the standard \orb\ system, the weights of the reprojection errors in the motion-only BA are assigned based on the image pyramid level at which a feature is extracted (1.0 for the original image resolution, lower weight for higher pyramid levels with lower resolutions).
\orb\ uses eight pyramid levels where the image resolution is reduced by a factor of 1.2 for each level.
In contrast, we only extract FAST corners at the original image resolution in \orbf\ and \orbfu.
\orb\ employs an outlier rejection strategy during bundle adjustment that is based on a $\chi^2$ test with significance level equal to 0.05. 
The $\chi^2$ test is executed 4 times, each time following 10 iterations of the optimizer. 
Points that do not pass the test are labeled as outliers and removed from subsequent optimizer iterations.
In \orbf, all reprojection error terms are assigned a uniform weight of 1.0.
In \orbfu, we utilize the confidence maps predicted by the flow network to weight the reprojection errors in the BA. 
This allows the backend to down-weight correspondences in regions where the network is uncertain.
In all versions, the BA employs the Huber loss function and the $\chi^2$ test. 

\comment{Because the network jointly estimates the optical flow and its corresponding uncertainty, the uncertainty formulation is inherently tied to the learned data association space. 
Consequently, combining classical ORB-based matches with this learned uncertainty is methodologically inconsistent.}
\section{Experiments}\label{sec:Experiments}
We evaluate the performance of \orb, \orbf, \orbfu, \droidvo, \droidslam, \dpvo, and \dpvslam, as detailed in Sec.~\ref {sec:Methodology}. 
These methods are tested on the \tartanair\ and \uzhfpv\ datasets, which serve as standard benchmarks for \vslam\ due to their inclusion of aggressive motion, large optical flow, and varying illumination conditions.
The primary objective of our evaluation is to compare the accuracy of the predicted camera trajectories. 
To this end, we employ the Absolute Trajectory Error (\ate) in translation (\atet\ [m]) and rotation (\ater\ [deg]) as the primary evaluation metrics~\cite{Zhang18iros}. 
These metrics are computed after aligning the estimated trajectories to the ground truth using a 7-DoF similarity transformation determined via the Umeyama method~\cite{umeyama2002least}.
The parameters for each \vslam\ system were specifically tuned for each dataset.
We run each \vslam\ system three times and report the average \ate.
Unless otherwise stated, each system processes every frame in the sequences, and the reported trajectory errors are averaged across all frames.
The experiments are run on a computer running Ubuntu 22.04, equipped with an Intel Core i9 2.3GHz CPU and an Nvidia A100 80GB GPU.

\begin{figure}[t!]
    \centering
    \includegraphics[width=1.0\linewidth]{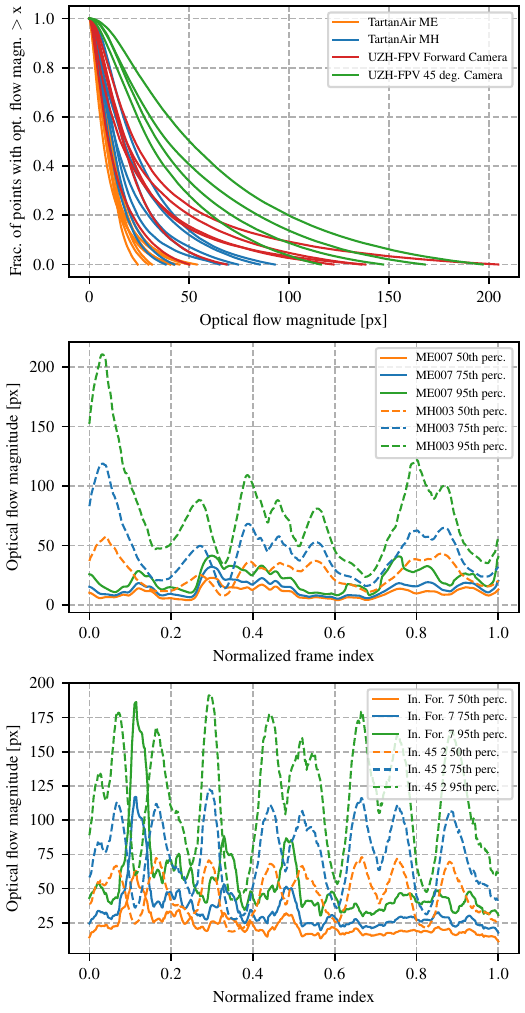}
    \vspace{-1mm}
    \caption{Analysis of optical flow magnitude between consecutive frames. \uzhfpv\ sequences exhibit higher flow magnitudes compared to \tartanair\ sequences. Top: Fraction of pixels with optical flow magnitude exceeding a threshold x. Middle: Comparison of the 50th, 75th, and 95th percentiles for a representative easy sequence (\textit{ME007}) and a hard sequence (\textit{MH003}) of the \tartanair\ dataset. Bottom: Similar percentile comparison for a representative forward camera sequence (\textit{Indoor Forward 7}) and a 45~deg camera sequence (\textit{Indoor 45 3}) of the \uzhfpv\ dataset.}
    \label{fig:cumulative_and_percentile_optflow}
\end{figure}
\begin{figure}[t!]
    \centering
    \includegraphics[width=1.0\linewidth]{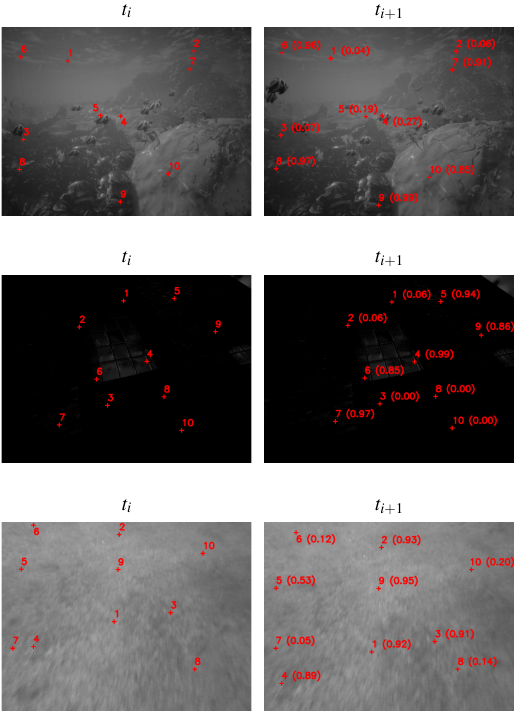}
    \vspace{-1mm}
    \caption{Optical flow visualization with associated confidence scores (in parentheses). Top, middle, and bottom rows show two consecutive images from the sequences \textit{ME004} and \textit{MH000} of the \tartanair\ dataset and the sequence \textit{In. 45 14} of the \uzhfpv\ dataset, respectively. Five high-confidence and five low-confidence predictions are visualized for each. The model predicts optical flow with high confidence even in textureless regions where visual tracking is difficult.}
    \label{fig:optflow_samples}
\end{figure}

\subsection{\tartanair\ Dataset}\label{sec:Experiments_TartanAir}
\subsubsection{Experimental Setup}
The \tartanair\ dataset~\cite{wang2020tartanair} comprises a diverse array of simulated environments, spanning indoor and outdoor settings such as urban, rural, domestic, public, and sci-fi locations. 
Each sequence is generated by defining a ground-truth trajectory and rendering corresponding RGB frames and poses, alongside multimodal data including depth, optical flow, events, LiDAR, and IMU measurements sampled along the trajectory.
Due to its scale, comprising hundreds of thousands of frames, the \tartanair\ dataset has been used to train the state-of-the-art DL-based \vslam\ systems used in this work (\droidslam, \dpvo, \dpvslam) and has become a standard dataset to train DL end-to-end 3D geometry reconstruction methods~\cite{leroy2024grounding, keetha2025mapanything}.
In this work, we use the 15 test sequences commonly used in related work to benchmark \vslam\ systems~\footnote{We excluded sequence \textit{MH004} from our evaluation due to rendering artifacts, such as missing surfaces.}. 
Categorized by difficulty into \textit{Easy} (\textit{ME}) and \textit{Hard} (\textit{MH}), these test sequences present significant challenges.
These include large optical flow displacements (predominantly in \textit{MH} sequences, as shown in the top plot of Fig.~\ref{fig:cumulative_and_percentile_optflow}), dynamic objects (\textit{ME004}), abrupt illumination changes (\textit{ME007}), and near-total darkness (\textit{MH000}, as shown in the middle row of Fig.~\ref{fig:optflow_samples}).

\subsubsection{Results}
The absolute trajectory errors for all test sequences are summarized in Table~\ref{tab:tartanair}. 
Additionally, Fig.~\ref{fig:traj_topview_viz_tartanair} provides a top-down visualization of ground-truth trajectories compared against estimates from \orbfu\ and the best-performing deep learning DL-based SLAM method according to the \atet.

The \orb\ variants integrated with learned optical flow (with or without uncertainty) significantly outperform the standard version of \orb. 
This performance gap is most pronounced in the \textit{Hard} (\textit{MH}) sequences, where standard \orb\ completes only two of the seven sequences. 
These sequences are characterized by large frame-to-frame optical flow displacements and adverse illumination, both of which degrade the reliability of ORB descriptor-based matching.
In the standard \orb\ algorithm, feature matching between consecutive frames relies on a constant velocity motion model to obtain an initial prediction of the location of features in the current frame. 
To maintain computational efficiency, the search for descriptor matches is restricted to a local region around these predicted coordinates.
Exhaustive search for all descriptors on the current frame is computationally intractable in a \vslam\ system.
However, the large accelerations and resulting high optical flow displacements present in the \textit{MH} sequences frequently violate the constant velocity assumption. 
This leads to inaccurate priors, causing the search windows to miss the true feature correspondences and resulting in tracking failure. 
The top plot of Fig.~\ref{fig:cumulative_and_percentile_optflow} confirms this, showing that the average optical flow magnitude in the \textit{MH} sequences (blue) consistently exceeds that of the (\textit{ME}) sequences (orange).
A comparison between \textit{ME007} (Easy) and \textit{MH003} (Hard) in the middle plot of Fig.~\ref{fig:cumulative_and_percentile_optflow} further illustrates this point. 
While standard \orb\ performs on par with the learned versions in \textit{ME007}, it suffers a complete loss of tracking in \textit{MH003}.
\textit{Hard} sequences are also characterized by challenging illumination conditions. 
Fig.~\ref{fig:optflow_samples} shows a set of optical flow-based feature matches and prediction confidence in a well-illuminated environment (sequence \textit{ME004} in the top plot) and in a low-illuminated and textureless region (sequence \textit{MH001} in the middle plot). 
The optical flow network maintains high-confidence predictions even in textureless or dark regions, where traditional descriptor matching fails. 
Consequently, \orbfu\ achieves a 55\% improvement in both \atet\ and \ater\ over standard \orb\ in sequence \textit{MH001}.

\begin{figure}[t!]
    \centering
    \includegraphics[width=1.0\linewidth]{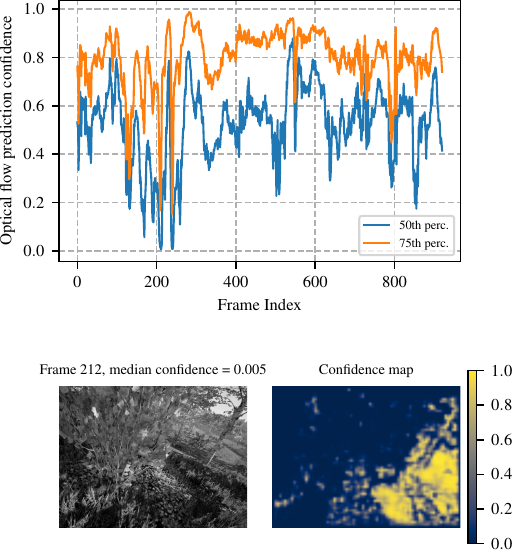}
    \vspace{-1mm}
    \caption{Optical flow prediction confidence analysis for \tartanair\ sequence \textit{MH005}. (Top) 50th and 75th percentiles of confidence across consecutive frames. (Bottom) Image and corresponding confidence map for the frame with the lowest median confidence score.}
    \label{fig:weight_distribution_and_zoomin_images_tartanair}
\end{figure}

The performance of \orbf\ and \orbfu\ is generally comparable, though incorporating flow uncertainty yields measurable gains in specific scenarios. 
For instance, in sequence \textit{ME004}, an underwater environment containing dynamic entities such as fish and moving vegetation, \orbfu\ improves the \atet\ by 33\% over \orbf. 
Similarly, in \textit{MH005}, which contains repetitive textures from dense vegetation, the flow uncertainty increases significantly (see Fig.~\ref{fig:weight_distribution_and_zoomin_images_tartanair}). In this situation, accounting for uncertainty leads to a 55\% improvement in \atet.

Overall, the integration of learned optical flow enables \orb\ to successfully process all sequences, bringing its performance close to state-of-the-art DL-based systems. 
The most significant performance gap between \orbfu\ and \dpvo\ is in sequence \textit{ME004} (1.20 vs. 0.42 \atet). 
While neither \orb\ nor \dpvo\ is explicitly designed for dynamic environments, this result suggests that \dpvo\ is more resilient in these situations. 
Finally, we observed that \dpvslam\ performance was primarily degraded by incorrect loop closure detections in these environments.
The average time required to process a single frame, starting with image loading and concluding with pose and map point prediction, is shown in the first row of Table~\ref{tab:runtime}.

\begin{table*}[t!]
\caption{Absolute Trajectory Error, in translation (\atet) and rotation (\ater), on the \uzhfpv\ dataset. Lower values indicate better performance, \textbf{bold} entries denote the best results in translation and rotation. The symbol '-' denotes that the SLAM system failed to maintain tracking for the entire sequence, and the symbol $\cdot^{*}$ denotes that the error metric is not computed for all the images included in the sequence. \textcolor{textblue}{Learned uncertainty is necessary to complete all the sequences.} \textcolor{textorange}{The performance of the \orb\ version with learned flow and uncertainty is better than that of the DL systems}.}
\vspace{3pt}
\label{tab:uzhfpv}
\setlength{\tabcolsep}{8pt}
\begin{tabularx}{\textwidth}{C|C|C|C|C|C|C|C}
\toprule 
\multirow[c]{1}{=}[-12pt]{\centering Sequence} & \multicolumn{7}{c}{Evaluation Metric: \atet [\unit{m}] / \ater [\unit{deg}]}  \\[4pt]
& \orb & \orbfTable & \orbfuTable & \droidvo & \droidslam & \dpvo & \dpvslam \\ [4pt]
 \midrule
 In. Forw. 3 & 0.09 / 0.45 & 0.12 / 0.44 & 0.07 / 0.42 & 1.55 / 0.88 & 0.79 / 2.25 & 0.40 / 1.11 & \textbf{0.02} / \textbf{0.37}\\
 In. Forw. 6 & 0.12 / 2.24 & 0.05 / 0.56 & 0.07 / 0.59 & 1.32 / 0.98 & 0.43 / 2.32 & 0.41 / 0.58 & \textbf{0.03} / \textbf{0.50}\\
 In. Forw. 7 & 0.27 / 3.40 & 0.54 / \textbf{1.11} & \textbf{0.20} / 2.60 & 5.11 / 21.67 & 5.11 / 22.64 & 0.63 / 2.31 & 0.65 / 1.86\\
 In. 45 2 & - & - & \textcolor{textblue}{0.24 / 2.13} & 1.09 / 6.93 & \textbf{0.06} / \textbf{1.61} & 1.06 / 7.42 & 0.92 / 6.67\\
 $\text{In. 45 9}^{*}$ & - & - & \textcolor{textblue}{0.32 / 6.11} & 0.90 / 4.39 & \textbf{0.15 }/ \textbf{2.12} & 0.81 / 4.45 & 0.71 / 4.48\\
 $\text{In. 45 14}^{*}$ & - & - & \textcolor{textblue}{\textbf{0.64} / \textbf{3.90}} & 5.27 / 22.41 & 5.26 / 22.05 & 2.79 / 9.24 & 3.46 / 10.09\\
 Out. Forw. 1 & 1.50 / 1.69 & 0.31 / 1.00 & 0.78 / 1.67 & 3.23 / 0.54 & 4.29 / 2.27 & 1.04 / 0.63 & \textbf{0.29} / \textbf{0.41}\\
 Out. Forw. 3 & - & \textbf{2.01} / 2.17 & 2.32 / 2.03 & 8.47 / \textbf{0.87} & 8.35 / 1.54 & 3.94 / 1.23 & 2.85 / 0.94\\
 Out. Forw. 5 & - & - & \textcolor{textblue}{\textbf{0.72} / 1.43} & 2.28 / 2.59 & 2.34 / 2.63 & 0.99 / \textbf{0.60} & 0.77 / 0.66\\
 Out. 45 1 & - & - & \textcolor{textblue}{\textbf{0.98} / \textbf{2.91}} & 1.91 / 5.17 & 1.91 / 5.26 & 1.69 / 3.27 & 1.06 / 3.44\\
 \midrule
 Average & - & - & \textcolor{textorange}{\textbf{0.63} / \textbf{2.38}} & \textcolor{textorange}{3.11 / 6.64} & \textcolor{textorange}{2.87 / 6.47} & \textcolor{textorange}{1.38 / 3.08} & \textcolor{textorange}{1.08 / 2.94}\\
\bottomrule
\end{tabularx}
\end{table*}
\begin{figure*}[t!]
    \centering
    \includegraphics[width=1.0\linewidth]{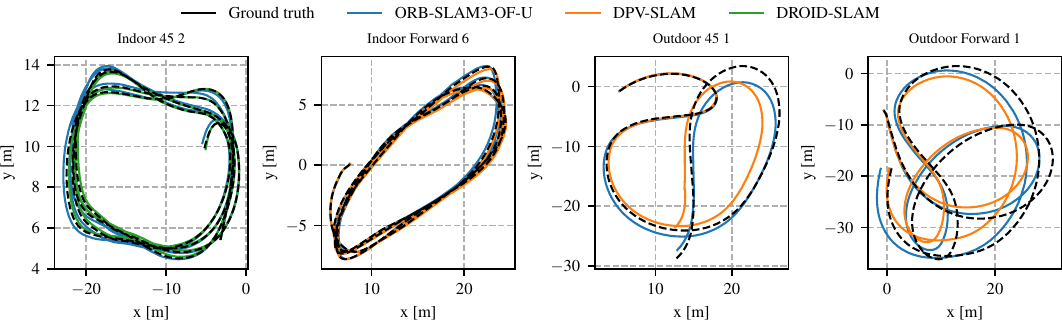}
    \vspace{-1mm}
    \caption{Top-down trajectories for four sequences from the \uzhfpv\ dataset. Ground truth is compared against trajectories from \orbfu\ and the best-performing DL-based SLAM method according to \atet. All predicted trajectories are aligned to the ground truth using the method~\cite{umeyama2002least} based on the first 100 frames of each sequence.}
    \label{fig:traj_topview_viz_uzhfpv}
\end{figure*}

\subsection{\uzhfpv\ Dataset}\label{sec:Experiments_UZHFPV}
\subsubsection{Experimental Setup}
The \uzhfpv\ dataset~\cite{delmerico2019we} consists of real-world indoor and outdoor sequences captured by a racing drone performing aggressive maneuvers. 
The drone is equipped with a synchronized camera and IMU, and was flown by an expert human pilot. 
The camera images are recorded at 30~Hz.
Sequences are categorized by their difficulty level (easy, medium, and hard) and camera orientation (forward-looking or 45-degree downward-looking).
The difficulty classification is primarily based on the magnitude of the optical flow. 
The \uzhfpv\ dataset is recognized as one of the most demanding \vslam\ benchmarks. 
Its aggressive camera motion induces optical flow magnitudes that significantly exceed those of established datasets such as TUM-RGBD~\cite{sturm2012benchmark}, EuRoC~\cite{burri2016euroc}, and \tartanair.
As illustrated in the top plot of Fig.~\ref{fig:cumulative_and_percentile_optflow}, the optical flow magnitude in the \uzhfpv\ sequences significantly exceeds that of the \tartanair\ dataset.
For our experiments, we selected a representative subset of sequences across different environments and camera configurations. 
Specifically, for the indoor (45-degree and forward) and outdoor (forward) configurations, we evaluate one easy sequence (\textit{In. 45 2}, \textit{In. Forw. 3}, \textit{Out. Forw. 1}), one medium sequence (\textit{In. 45 9}, \textit{In. Forw. 6}, \textit{Out. Forw. 3}), and one hard sequence (\textit{In. 45 14}, \textit{In. Forw. 7}, \textit{Out. Forw. 5}). 
Additionally, we include \textit{Out. 45 1}, which is the only sequence with publicly available ground truth for the outdoor 45-degree configuration.
The ground truth for this dataset was computed via a joint optimization framework~\cite{cioffi2022continuous} that minimizes visual, inertial, and 3D position residuals. 
The 3D position constraints were provided by a robotic total station tracking a prism mounted on the drone, ensuring high-precision reference trajectories even during high-speed flight.

\subsubsection{Results}
The absolute trajectory errors for the sequences evaluated in our experiments are summarized in Table~\ref{tab:uzhfpv}. 
Correspondingly, Fig.~\ref{fig:traj_topview_viz_uzhfpv} provides a top-down visualization of ground-truth trajectories alongside estimates from \orbfu\ and the best-performing DL-based SLAM method according to the \atet.

Consistent with our findings on the \tartanair\ dataset, the integration of learned optical flow and uncertainty significantly enhances performance over standard \orb. 
This gap is most pronounced in configurations where the camera is mounted at a 45~deg downward angle. 
In these sequences, standard \orb\ fails to complete any of the trajectories. 
In contrast, for indoor sequences with a forward-facing camera, the performance gains of \orbfu\ are more modest. 
The 45~deg configuration is particularly challenging due to large frame-to-frame optical flow displacements (as shown in the top and bottom plots of Fig.~\ref{fig:cumulative_and_percentile_optflow}) and the presence of textureless regions when the camera directly faces the floor (as shown in the bottom plot of Fig.~\ref{fig:optflow_samples}).

\begin{figure}[t!]
    \centering
    \includegraphics[width=1.0\linewidth]{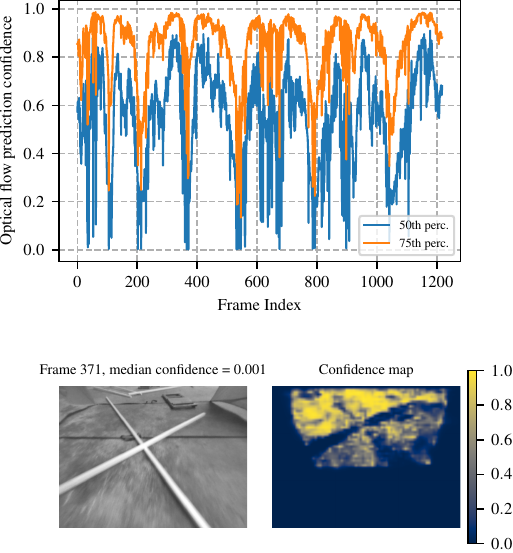}
    \vspace{-1mm}
    \caption{Optical flow prediction confidence analysis for \uzhfpv\ sequence \textit{In. 45 2}. (Top) 50th and 75th percentiles of confidence across consecutive frames. (Bottom) Image and corresponding confidence map for the frame with the lowest median confidence score.}
    \label{fig:weight_distribution_and_zoomin_images_uzhfpv}
\end{figure}

Notably, the performance difference between \orbfu\ and \orbf\ is much more significant here than in the \tartanair\ experiments. 
Specifically, \orbf\ fails to complete any of the 45~deg sequences as well as the \textit{Out. Forw. 5} (hard outdoor) sequence. 
In the 45~deg orientation, the drone’s high velocity relative to its proximity to the floor results in very low optical flow confidence (illustrated in Fig.~\ref{fig:weight_distribution_and_zoomin_images_uzhfpv}).
Additionally, the rapid motion causes large portions of the scene to exit the field of view between successive frames, drastically reducing visual overlap.

In sequences \textit{In. 45 9} and \textit{In. 45 14}, tracking failures occurred for all versions of \orb\ due to a hardware-related issue: two missing recorded frames created a 100~ms temporal gap. 
The resulting lack of visual overlap caused the systems to lose the tracking of the current frame. 
Consequently, for these sequences, the error metrics for all the systems are computed only for the first 76\% and 84\% of the frames, respectively. 
DL-based methods maintained tracking continuity by initializing frame poses via a constant velocity motion model. 
In the absence of visual measurements to constrain the local bundle adjustment, these methods simply propagate the prior motion model, whereas the \orb-based systems trigger a tracking-lost state.
Overall, the integration of learned optical flow and uncertainty allows \orb\ to surpass the performance of state-of-the-art DL-based systems on this challenging dataset.
The average time required to process a single frame is shown in the second row of Table~\ref{tab:runtime}.

\begin{table*}[t!]
\caption{Runtime evaluation (in seconds) in two selected sequences from the \tartanair\ and \uzhfpv\ datasets. Reported values represent the average time required to process a single frame, starting with image loading and concluding with pose and map point prediction.}
\vspace{3pt}
\label{tab:runtime}
\setlength{\tabcolsep}{8pt}
\begin{tabularx}{\textwidth}{C|C|C|C|C|C|C|C}
\toprule 
\multirow[c]{1}{=}[-12pt]{\centering Sequence} & \multicolumn{7}{c}{Average time per frame [\unit{sec}]}  \\[4pt]
& \orb & \orbfTable & \orbfuTable & \droidvo & \droidslam & \dpvo & \dpvslam \\ [4pt]
 \midrule
 \tartanair\ ME007 & 0.024 & 0.050 & 0.050 & 0.101 & 0.201 & 0.043 & 0.048 \\
 &  &  &  &  &  &  &  \\
 \uzhfpv\ Ind. Forw. 3 & 0.025 & 0.051 & 0.051 & 0.159 & 0.360 & 0.038 & 0.091 \\
\bottomrule
\end{tabularx}
\end{table*}
\section{Discussion}\label{sec:Discussion}
\subsubsection{\qf}
The results of our study provide strong empirical evidence that learned 2D data association is indeed the primary reason for the superior performance of modern DL-based \vslam\ systems.
While the baseline \orb\ is consistently outperformed by DL-based methods, the integration of learned optical flow brings the performance of \orb\ to a level comparable to state-of-the-art DL systems, regardless of the underlying architectural differences. 
%

The advantages of learned 2D data association are most prominent in challenging scenarios: high optical flow displacements caused by aggressive motion (as observed in the \uzhfpv\ dataset) and adverse illumination or low-texture environments (as seen in the \textit{MH} sequences of the \tartanair\ dataset). 
In these cases, classical descriptor-matching algorithms often fail or require their parameters to be manually adapted to the specific environment. 
Our study demonstrates that the learned frontend effectively addresses these challenges, suggesting that next-generation \vslam\ systems should prioritize learned modules for data association.

Furthermore, DL-based frontends alleviate the burden of manual parameter tuning. 
Classical geometry-based pipelines rely on numerous heuristics, such as pixel-intensity thresholds for feature extraction and Hamming distance limits for descriptor matching. 
Identifying a universal parameter set that generalizes across diverse environments is notoriously challenging and typically requires expert users. 
By contrast, DL techniques shift the challenge from manual tuning to the collection of diverse and representative training data. 
Provided that the training data reflects the expected deployment conditions, the learned frontend generalizes more robustly. 
This direction, also explored in recent works such as~\cite{messikommer2024reinforcement, nascivera2026image}, represents a promising path for making \vslam\ more robust in visually degraded scenarios.

\subsubsection{\qs}
Our evaluation suggests that learned uncertainty estimation is crucial for robust localization, particularly when the underlying 2D data association process is noisy. 
This is most evident in the \uzhfpv\ dataset, where the domain shift from the simulated \tartanair\ training data to high-velocity, real-world drone racing images resulted in increased optical flow prediction uncertainty. 
In these high-noise regimes, classical outlier rejection strategies, such as $\chi^2$ statistical tests and robust cost functions (e.g., the Huber norm) utilized in \orb, often fail to sufficiently filter or down-weight erroneous correspondences during bundle adjustment.
Our study empirically demonstrates that incorporating learned uncertainty allows bundle adjustment to account for the inherent stochasticity of the 2D data associations. 
By weighting the reprojection error according to the network's confidence, the system achieves a level of robustness that is difficult to replicate with fixed statistical thresholds. 
Consequently, we conclude that learned data association coupled with learned uncertainty is necessary for robust \vslam\ systems operating in visually challenging environments.

\subsubsection{\qt}
\comment{A key insight from our evaluation is that the recurrent architecture utilized by \droidslam, \dpvo, and \dpvslam\ is not strictly necessary to achieve state-of-the-art trajectory estimation accuracy. 
On the UZH-FPV dataset, our learned variant of \orb\ achieves the best performance by leveraging the classical \orb\ feed-forward architecture combined with learned 2D data association and uncertainty estimation.
This superior performance stems from the fact that the classical \orb\ system is a sophisticated, multi-stage pipeline featuring frame-to-map registration, local mapping, complex keyframe selection, and loop closure, where each block is governed by numerous interdependent parameters.
Crucially, this architecture incorporates structural redundancies that enable accurate trajectory estimation even in the presence of a significant number of outlier measurements, such as noisy 2D data associations.
This suggests that the fundamental driver behind the performance of modern deep-learning-based \vslam\ systems is primarily the robustness of learned 2D data association and its corresponding uncertainty. 
This takeaway provides a clear design principle for next-generation SLAM: practitioners can achieve state-of-the-art accuracy and robustness by combining learned frontends with interpretable and computationally efficient geometric backends.}
\section{Conclusion}\label{sec:Conclusion}
In this work, we present a systematic study of the architectural drivers behind the superior performance of DL-based \vslam\ systems relative to classical geometry-based frameworks.
By utilizing \orb\ as a classical baseline and integrating the optical flow network of \droidslam, we addressed the fundamental research question: \question\
Our empirical evaluation on the \tartanair\ and \uzhfpv\ datasets demonstrates that learned 2D data association is indeed the primary factor enabling DL-based systems to outperform classical geometry-based pipelines. 
The integration of learned flow allows \orb\ to overcome traditional failure modes, such as large frame-to-frame displacements, motion blur, and challenging illumination conditions, that typically cause localization failures.
Our results show that when equipped with learned data association, a classical SLAM architecture like \orb\ achieves state-of-the-art performance.
Furthermore, our study confirms that learned uncertainty estimation is indispensable for robust localization.
While classical outlier rejection methods perform well in many cases, they are often insufficient in high-noise situations. 
We found that weighting reprojection errors with learned confidence significantly improves trajectory accuracy in degraded visual conditions compared to relying on classical outlier rejection methods.
In summary, our findings suggest that next-generation \vslam\ architectures should integrate the robustness of learned frontends for data association and uncertainty estimation with geometric bundle adjustment in classical feed-forward architectures.

{\bibliographystyle{unsrtnat}
\bibliography{references}}

\begin{thebibliography}{56}
\providecommand{\natexlab}[1]{#1}
\providecommand{\url}[1]{\texttt{#1}}
\expandafter\ifx\csname urlstyle\endcsname\relax
  \providecommand{\doi}[1]{doi: #1}\else
  \providecommand{\doi}{doi: \begingroup \urlstyle{rm}\Url}\fi

\bibitem[Ebadi et~al.(2024)Ebadi, Bernreiter, Biggie, et~al.]{ebadi2024present}
Kamak Ebadi, Lukas Bernreiter, Harel Biggie, et~al.
\newblock Present and future of {SLAM} in extreme environments: The {DARPA SubT} challenge.
\newblock \emph{{IEEE} Trans. Robot.}, 2024.

\bibitem[Janai et~al.(2020)Janai, G{\"u}ney, Behl, and Geiger]{janai2020computer}
Joel Janai, Fatma G{\"u}ney, Aseem Behl, and Andreas Geiger.
\newblock Computer vision for autonomous vehicles: Problems, datasets and state of the art.
\newblock \emph{Foundations and Trends in Computer Graphics and Vision}, 2020.

\bibitem[Engel et~al.(2023)Engel, Somasundaram, Goesele, et~al.]{engel2023project}
Jakob Engel, Kiran Somasundaram, Michael Goesele, et~al.
\newblock Project aria: A new tool for egocentric multi-modal ai research.
\newblock \emph{arXiv preprint arXiv:2308.13561}, 2023.
\newblock Accessed 2026-04-20.

\bibitem[Pfreundschuh et~al.(2025)Pfreundschuh, Cioffi, von Einem, et~al.]{pfreundschuh2025sight}
Patrick Pfreundschuh, Giovanni Cioffi, Cornelius von Einem, et~al.
\newblock Sight guide demonstrates robotics-inspired vision assistance at the cybathlon.
\newblock \emph{Science Robotics}, 2025.

\bibitem[Most et~al.(2025)Most, Hein, Giraud, et~al.]{most2025acquiring}
Emma Most, Jonas Hein, Fr{\'e}d{\'e}ric Giraud, et~al.
\newblock Acquiring submillimeter-accurate multi-task vision datasets for computer-assisted orthopedic surgery.
\newblock \emph{Int. Journal of Computer Assisted Radiology and Surgery}, 2025.

\bibitem[Carlone et~al.(2026)Carlone, Kim, Barfoot, Cremers, and Dellaert]{slam-handbook}
Luca Carlone, Ayoung Kim, Timothy Barfoot, Daniel Cremers, and Frank Dellaert, editors.
\newblock \emph{{SLAM Handbook.} From Localization and Mapping to Spatial Intelligence}.
\newblock Cambridge University Press, 2026.

\bibitem[Wang et~al.(2020)Wang, Zhu, Wang, Hu, Qiu, Wang, Hu, Kapoor, and Scherer]{wang2020tartanair}
Wenshan Wang, Delong Zhu, Xiangwei Wang, Yaoyu Hu, Yuheng Qiu, Chen Wang, Yafei Hu, Ashish Kapoor, and Sebastian Scherer.
\newblock Tartanair: A dataset to push the limits of visual slam.
\newblock In \emph{IEEE/RSJ Int. Conf. Intell. Robot. Syst. (IROS)}, 2020.

\bibitem[Delmerico et~al.(2019)Delmerico, Cieslewski, Rebecq, Faessler, and Scaramuzza]{delmerico2019we}
Jeffrey Delmerico, Titus Cieslewski, Henri Rebecq, Matthias Faessler, and Davide Scaramuzza.
\newblock Are we ready for autonomous drone racing? the uzh-fpv drone racing dataset.
\newblock In \emph{{IEEE} Int. Conf. Robot. Autom. (ICRA)}, 2019.

\bibitem[Gehrig et~al.(2021)Gehrig, Aarents, Gehrig, and Scaramuzza]{gehrig2021dsec}
Mathias Gehrig, Willem Aarents, Daniel Gehrig, and Davide Scaramuzza.
\newblock Dsec: A stereo event camera dataset for driving scenarios.
\newblock \emph{{IEEE} Robot. Autom. Lett.}, 2021.

\bibitem[Helmberger et~al.(2022)Helmberger, Morin, Berner, Kumar, Cioffi, and Scaramuzza]{helmberger2022hilti}
Michael Helmberger, Kristian Morin, Beda Berner, Nitish Kumar, Giovanni Cioffi, and Davide Scaramuzza.
\newblock The hilti slam challenge dataset.
\newblock \emph{{IEEE} Robot. Autom. Lett.}, 2022.

\bibitem[Palazzolo et~al.(2019)Palazzolo, Behley, Lottes, Gigu\`ere, and Stachniss]{palazzolo2019iros}
E.~Palazzolo, J.~Behley, P.~Lottes, P.~Gigu\`ere, and C.~Stachniss.
\newblock {ReFusion: 3D Reconstruction in Dynamic Environments for RGB-D Cameras Exploiting Residuals}.
\newblock In \emph{IEEE/RSJ Int. Conf. Intell. Robot. Syst. (IROS)}, 2019.

\bibitem[Wang et~al.(2017)Wang, Clark, Wen, and Trigoni]{wang2017deepvo}
Sen Wang, Ronald Clark, Hongkai Wen, and Niki Trigoni.
\newblock Deepvo: Towards end-to-end visual odometry with deep recurrent convolutional neural networks.
\newblock In \emph{{IEEE} Int. Conf. Robot. Autom. (ICRA)}, 2017.

\bibitem[Wang et~al.(2021)Wang, Hu, and Scherer]{wang2021tartanvo}
Wenshan Wang, Yaoyu Hu, and Sebastian Scherer.
\newblock Tartanvo: A generalizable learning-based vo.
\newblock In \emph{Conf. on Robot. Learning (CoRL)}, 2021.

\bibitem[Teed and Deng(2021)]{teed2021droid}
Zachary Teed and Jia Deng.
\newblock Droid-slam: Deep visual slam for monocular, stereo, and rgb-d cameras.
\newblock \emph{Advances in Neural Information Processing Systems}, 34:\penalty0 16558--16569, 2021.

\bibitem[Teed et~al.(2022)Teed, Lipson, and Deng]{teed2022deep}
Zachary Teed, Lahav Lipson, and Jia Deng.
\newblock Deep patch visual odometry.
\newblock \emph{arXiv preprint arXiv:2208.04726}, 2022.
\newblock URL \url{https://arxiv.org/abs/2208.04726}.
\newblock Retrieved on Jan. 4th, 2023.

\bibitem[Lipson et~al.(2024)Lipson, Teed, and Deng]{lipson2024deep}
Lahav Lipson, Zachary Teed, and Jia Deng.
\newblock Deep patch visual slam.
\newblock In \emph{Eur. Conf. Comput. Vis. (ECCV)}, 2024.

\bibitem[Teed and Deng(2020)]{teed2020raft}
Zachary Teed and Jia Deng.
\newblock Raft: Recurrent all-pairs field transforms for optical flow.
\newblock In \emph{Eur. Conf. Comput. Vis. (ECCV)}, 2020.

\bibitem[Zhu et~al.(2022)Zhu, Peng, Larsson, Xu, Bao, Cui, Oswald, and Pollefeys]{zhu2022nice}
Zihan Zhu, Songyou Peng, Viktor Larsson, Weiwei Xu, Hujun Bao, Zhaopeng Cui, Martin~R Oswald, and Marc Pollefeys.
\newblock Nice-slam: Neural implicit scalable encoding for slam.
\newblock In \emph{{IEEE} Conf. Comput. Vis. Pattern Recog. (CVPR)}, 2022.

\bibitem[Zhu et~al.(2024)Zhu, Peng, Larsson, Cui, Oswald, Geiger, and Pollefeys]{zhu2024nicer}
Zihan Zhu, Songyou Peng, Viktor Larsson, Zhaopeng Cui, Martin~R Oswald, Andreas Geiger, and Marc Pollefeys.
\newblock Nicer-slam: Neural implicit scene encoding for rgb slam.
\newblock In \emph{Int. Conf. on 3D Vision (3DV)}, 2024.

\bibitem[Matsuki et~al.(2024)Matsuki, Murai, Kelly, and Davison]{matsuki2024gaussian}
Hidenobu Matsuki, Riku Murai, Paul~HJ Kelly, and Andrew~J Davison.
\newblock Gaussian splatting slam.
\newblock In \emph{{IEEE} Conf. Comput. Vis. Pattern Recog. (CVPR)}, 2024.

\bibitem[Yugay et~al.(2023)Yugay, Li, Gevers, and Oswald]{yugay2023gaussian}
Vladimir Yugay, Yue Li, Theo Gevers, and Martin~R Oswald.
\newblock Gaussian-slam: Photo-realistic dense slam with gaussian splatting.
\newblock \emph{arXiv preprint arXiv:2312.10070}, 2023.
\newblock Accessed 2026-05-06.

\bibitem[Murai et~al.(2025)Murai, Dexheimer, and Davison]{murai2025mast3r}
Riku Murai, Eric Dexheimer, and Andrew~J Davison.
\newblock Mast3r-slam: Real-time dense slam with 3d reconstruction priors.
\newblock In \emph{{IEEE} Conf. Comput. Vis. Pattern Recog. (CVPR)}, 2025.

\bibitem[Maggio et~al.(2025)Maggio, Lim, and Carlone]{maggio2025vggt-slam}
Dominic Maggio, Hyungtae Lim, and Luca Carlone.
\newblock Vggt-slam: Dense rgb slam optimized on the sl (4) manifold.
\newblock \emph{Conf. Neural Inf. Process. Syst. (NIPS)}, 2025.

\bibitem[Maggio and Carlone(2026)]{maggio2025vggt-slam2}
Dominic Maggio and Luca Carlone.
\newblock Vggt-slam 2.0: Real-time dense feed-forward scene reconstruction.
\newblock \emph{arXiv preprint arXiv:2601.19887}, 2026.
\newblock Accessed 2026-04-20.

\bibitem[Qiu et~al.(2025)Qiu, Chen, Zhang, Wang, and Scherer]{qiu2025mac}
Yuheng Qiu, Yutian Chen, Zihao Zhang, Wenshan Wang, and Sebastian Scherer.
\newblock Mac-vo: Metrics-aware covariance for learning-based stereo visual odometry.
\newblock In \emph{{IEEE} Int. Conf. Robot. Autom. (ICRA)}, 2025.

\bibitem[Zhao et~al.(2025)Zhao, Zhou, Zhang, Zhang, Wang, Wang, and Scherer]{zhao2025resilient}
Shibo Zhao, Sifan Zhou, Yuchen Zhang, Ji~Zhang, Chen Wang, Wenshan Wang, and Sebastian Scherer.
\newblock Resilient odometry via hierarchical adaptation.
\newblock \emph{Science Robotics}, 2025.

\bibitem[Dellaert(2012)]{dellaert2012factor}
Frank Dellaert.
\newblock Factor graphs and gtsam: A hands-on introduction.
\newblock \emph{Georgia Institute of Technology, Tech. Rep.}, 2012.

\bibitem[Sol{\`a} et~al.(2018)Sol{\`a}, Deray, and Atchuthan]{sola2018micro}
Joan Sol{\`a}, Jeremie Deray, and Dinesh Atchuthan.
\newblock A micro {L}ie theory for state estimation in robotics.
\newblock \emph{arXiv preprint arXiv:1812.01537}, 2018.
\newblock Accessed 2026-04-20.

\bibitem[Grisetti et~al.(2011)Grisetti, K{\"u}mmerle, Stachniss, and Burgard]{grisetti2011tutorial}
Giorgio Grisetti, Rainer K{\"u}mmerle, Cyrill Stachniss, and Wolfram Burgard.
\newblock A tutorial on graph-based slam.
\newblock \emph{IEEE Intel. Transportation Sys. Mag.}, 2011.

\bibitem[Mur-Artal et~al.(2015)Mur-Artal, Montiel, and Tardos]{mur2015orb}
Raul Mur-Artal, Jose Maria~Martinez Montiel, and Juan~D Tardos.
\newblock Orb-slam: A versatile and accurate monocular slam system.
\newblock \emph{{IEEE} Trans. Robot.}, 2015.

\bibitem[Mur-Artal and Tard{\'o}s(2017)]{mur2017orb}
Raul Mur-Artal and Juan~D Tard{\'o}s.
\newblock Orb-slam2: An open-source slam system for monocular, stereo, and rgb-d cameras.
\newblock \emph{{IEEE} Trans. Robot.}, 2017.

\bibitem[Campos et~al.(2021)Campos, Elvira, Rodr{\'\i}guez, Montiel, and Tard{\'o}s]{campos2021orb}
Carlos Campos, Richard Elvira, Juan J~G{\'o}mez Rodr{\'\i}guez, Jos{\'e}~MM Montiel, and Juan~D Tard{\'o}s.
\newblock Orb-slam3: An accurate open-source library for visual, visual--inertial, and multimap slam.
\newblock \emph{{IEEE} Trans. Robot.}, 2021.

\bibitem[Rosten and Drummond(2006)]{rosten2006machine}
Edward Rosten and Tom Drummond.
\newblock Machine learning for high-speed corner detection.
\newblock In \emph{Eur. Conf. Comput. Vis. (ECCV)}, 2006.

\bibitem[Rublee et~al.(2011)Rublee, Rabaud, Konolige, and Bradski]{rublee2011orb}
Ethan Rublee, Vincent Rabaud, Kurt Konolige, and Gary Bradski.
\newblock Orb: An efficient alternative to sift or surf.
\newblock In \emph{Int. Conf. Comput. Vis. (ICCV)}, 2011.

\bibitem[Engel et~al.(2017)Engel, Koltun, and Cremers]{engel2017direct}
Jakob Engel, Vladlen Koltun, and Daniel Cremers.
\newblock Direct sparse odometry.
\newblock \emph{{IEEE} Trans. Pattern Anal. Mach. Intell.}, 2017.

\bibitem[Forster et~al.(2016)Forster, Zhang, Gassner, Werlberger, and Scaramuzza]{forster2016svo}
Christian Forster, Zichao Zhang, Michael Gassner, Manuel Werlberger, and Davide Scaramuzza.
\newblock Svo: Semidirect visual odometry for monocular and multicamera systems.
\newblock \emph{{IEEE} Trans. Robot.}, 2016.

\bibitem[Davison et~al.(2007)Davison, Reid, Molton, and Stasse]{davison2007monoslam}
Andrew~J Davison, Ian~D Reid, Nicholas~D Molton, and Olivier Stasse.
\newblock Monoslam: Real-time single camera slam.
\newblock \emph{{IEEE} Trans. Pattern Anal. Mach. Intell.}, 2007.

\bibitem[Mourikis and Roumeliotis(2007)]{mourikis2007multi}
Anastasios~I Mourikis and Stergios~I Roumeliotis.
\newblock A multi-state constraint kalman filter for vision-aided inertial navigation.
\newblock In \emph{{IEEE} Int. Conf. Robot. Autom. (ICRA)}, 2007.

\bibitem[Leutenegger et~al.(2015)Leutenegger, Lynen, Bosse, Siegwart, and Furgale]{leutenegger2015keyframe}
Stefan Leutenegger, Simon Lynen, Michael Bosse, Roland Siegwart, and Paul Furgale.
\newblock Keyframe-based visual--inertial odometry using nonlinear optimization.
\newblock \emph{Int. J. Robot. Research}, 2015.

\bibitem[Qin et~al.(2018)Qin, Li, and Shen]{qin2018vins}
Tong Qin, Peiliang Li, and Shaojie Shen.
\newblock Vins-mono: A robust and versatile monocular visual-inertial state estimator.
\newblock \emph{{IEEE} Trans. Robot.}, 2018.

\bibitem[Strasdat et~al.(2012)Strasdat, Montiel, and Davison]{strasdat2012visual}
Hauke Strasdat, Jos{\'e}~MM Montiel, and Andrew~J Davison.
\newblock Visual slam: why filter?
\newblock \emph{Image and Vision Computing}, 2012.

\bibitem[Engel et~al.(2026)Engel, Tard{\'o}s, Civera, Chli, Leutenegger, Dellaert, and Cremers]{sh-ch7-vision}
Jakob Engel, Juan~D. Tard{\'o}s, Javier Civera, Margarita Chli, Stefan Leutenegger, Frank Dellaert, and Daniel Cremers.
\newblock Visual {SLAM}.
\newblock In Luca Carlone, Ayoung Kim, Timothy Barfoot, Daniel Cremers, and Frank Dellaert, editors, \emph{{SLAM Handbook.} From Localization and Mapping to Spatial Intelligence}. Cambridge University Press, 2026.

\bibitem[Yang et~al.(2020)Yang, Stumberg, Wang, and Cremers]{yang2020d3vo}
Nan Yang, Lukas~von Stumberg, Rui Wang, and Daniel Cremers.
\newblock D3vo: Deep depth, deep pose and deep uncertainty for monocular visual odometry.
\newblock In \emph{{IEEE} Conf. Comput. Vis. Pattern Recog. (CVPR)}, 2020.

\bibitem[Sucar et~al.(2021)Sucar, Liu, Ortiz, and Davison]{sucar2021imap}
Edgar Sucar, Shikun Liu, Joseph Ortiz, and Andrew~J Davison.
\newblock imap: Implicit mapping and positioning in real-time.
\newblock In \emph{Int. Conf. Comput. Vis. (ICCV)}, 2021.

\bibitem[Mildenhall et~al.(2021)Mildenhall, Srinivasan, Tancik, Barron, Ramamoorthi, and Ng]{mildenhall2021nerf}
Ben Mildenhall, Pratul~P Srinivasan, Matthew Tancik, Jonathan~T Barron, Ravi Ramamoorthi, and Ren Ng.
\newblock Nerf: Representing scenes as neural radiance fields for view synthesis.
\newblock \emph{Communications of the ACM}, 2021.

\bibitem[Leroy et~al.(2024)Leroy, Cabon, and Revaud]{leroy2024grounding}
Vincent Leroy, Yohann Cabon, and J{\'e}r{\^o}me Revaud.
\newblock Grounding image matching in 3d with mast3r.
\newblock In \emph{Eur. Conf. Comput. Vis. (ECCV)}, 2024.

\bibitem[Wang et~al.(2025)Wang, Chen, Karaev, Vedaldi, Rupprecht, and Novotny]{wang2025vggt}
Jianyuan Wang, Minghao Chen, Nikita Karaev, Andrea Vedaldi, Christian Rupprecht, and David Novotny.
\newblock Vggt: Visual geometry grounded transformer.
\newblock In \emph{{IEEE} Conf. Comput. Vis. Pattern Recog. (CVPR)}, 2025.

\bibitem[G{\'a}lvez-L{\'o}pez and Tardos(2012)]{galvez2012bags}
Dorian G{\'a}lvez-L{\'o}pez and Juan~D Tardos.
\newblock Bags of binary words for fast place recognition in image sequences.
\newblock \emph{{IEEE} Trans. Robot.}, 2012.

\bibitem[Umeyama(2002)]{umeyama2002least}
Shinji Umeyama.
\newblock Least-squares estimation of transformation parameters between two point patterns.
\newblock \emph{{IEEE} Trans. Pattern Anal. Mach. Intell.}, 2002.

\bibitem[Zhang and Scaramuzza(2018)]{Zhang18iros}
Zichao Zhang and Davide Scaramuzza.
\newblock A tutorial on quantitative trajectory evaluation for visual(-inertial) odometry.
\newblock In \emph{IEEE/RSJ Int. Conf. Intell. Robot. Syst. (IROS)}, 2018.

\bibitem[Keetha et~al.(2026)Keetha, M{\"u}ller, Sch{\"o}nberger, Porzi, Zhang, Fischer, Knapitsch, Zauss, Weber, Antunes, et~al.]{keetha2025mapanything}
Nikhil Keetha, Norman M{\"u}ller, Johannes Sch{\"o}nberger, Lorenzo Porzi, Yuchen Zhang, Tobias Fischer, Arno Knapitsch, Duncan Zauss, Ethan Weber, Nelson Antunes, et~al.
\newblock Mapanything: Universal feed-forward metric 3d reconstruction.
\newblock In \emph{Int. Conf. on 3D Vision (3DV)}, 2026.

\bibitem[Sturm et~al.(2012)Sturm, Engelhard, Endres, Burgard, and Cremers]{sturm2012benchmark}
J{\"u}rgen Sturm, Nikolas Engelhard, Felix Endres, Wolfram Burgard, and Daniel Cremers.
\newblock A benchmark for the evaluation of rgb-d slam systems.
\newblock In \emph{IEEE/RSJ Int. Conf. Intell. Robot. Syst. (IROS)}, 2012.

\bibitem[Burri et~al.(2016)Burri, Nikolic, Gohl, Schneider, Rehder, Omari, Achtelik, and Siegwart]{burri2016euroc}
Michael Burri, Janosch Nikolic, Pascal Gohl, Thomas Schneider, Joern Rehder, Sammy Omari, Markus~W Achtelik, and Roland Siegwart.
\newblock The euroc micro aerial vehicle datasets.
\newblock \emph{Int. J. Robot. Research}, 2016.

\bibitem[Cioffi et~al.(2022)Cioffi, Cieslewski, and Scaramuzza]{cioffi2022continuous}
Giovanni Cioffi, Titus Cieslewski, and Davide Scaramuzza.
\newblock Continuous-time vs. discrete-time vision-based slam: A comparative study.
\newblock \emph{{IEEE} Robot. Autom. Lett.}, 2022.

\bibitem[Messikommer et~al.(2024)Messikommer, Cioffi, Gehrig, and Scaramuzza]{messikommer2024reinforcement}
Nico Messikommer, Giovanni Cioffi, Mathias Gehrig, and Davide Scaramuzza.
\newblock Reinforcement learning meets visual odometry.
\newblock In \emph{Eur. Conf. Comput. Vis. (ECCV)}, 2024.

\bibitem[Nascivera et~al.(2026)Nascivera, Bauersfeld, Delaune, and Scaramuzza]{nascivera2026image}
Simone Nascivera, Leonard Bauersfeld, Jeff Delaune, and Davide Scaramuzza.
\newblock Image-conditioned adaptive parameter tuning for visual odometry frontends.
\newblock \emph{arXiv preprint arXiv:2603.21785}, 2026.
\newblock Accessed 2026-04-15.

\end{thebibliography}
\clearpage
\end{document}